\documentclass[conference]{IEEEtran}
\IEEEoverridecommandlockouts
\usepackage[utf8]{inputenc}
\usepackage{textcomp}
\usepackage{cite}
\usepackage{listings}

\usepackage{enumitem}

\usepackage{pgfplots}
\DeclareUnicodeCharacter{2212}{−}
\usepgfplotslibrary{groupplots,dateplot}
\usetikzlibrary{patterns,shapes.arrows}
\pgfplotsset{compat=newest}
\usepackage{booktabs}
\usepackage[bookmarks=false]{hyperref}
\usepackage[all]{hypcap}
\usepackage{url}
\usepackage{amsmath,amssymb,amsfonts}
\usepackage{siunitx}
\usepackage{graphicx}
\graphicspath{ {./Used_Images/} }
\def\BibTeX{{\rm B\kern-.05em{\sc i\kern-.025em b}\kern-.08em
    T\kern-.1667em\lower.7ex\hbox{E}\kern-.125emX}}
\usepackage{booktabs}
\usepackage{threeparttable}    

\begin{document}
\title{REMONI: An Autonomous System Integrating Wearables and Multimodal Large Language Models for Enhanced Remote Health Monitoring %
}

\author{%
\IEEEauthorblockN{%
Thanh Cong Ho\IEEEauthorrefmark{1}, Farah Kharrat\IEEEauthorrefmark{1}, Abderrazek Abid\IEEEauthorrefmark{1}, Fakhri Karray\IEEEauthorrefmark{1}\IEEEauthorrefmark{2} }
\IEEEauthorblockA{\IEEEauthorrefmark{1}Mohamed bin Zayed University of Artificial Intelligence\\
Masdar City, Abu Dhabi, UAE\\
Email: \{Thanh.Ho, Farah.Kharrat, Abid.Abderrazek, Fakhri.Karray\}@mbzuai.ac.ae}
\IEEEauthorblockA{\IEEEauthorrefmark{2}Department of Electrical and Computer Engineering\\
University of Waterloo, Waterloo, ON, Canada N2L 3G1\\
Email: karray@uwaterloo.ca}
\IEEEauthorblockA{\IEEEauthorrefmark{3}College of Computer and Information Sciences\\
 Prince Sultan University\\
Email: akoubaa@psu.edu.sa}%
}

\maketitle

\begin{abstract}
With the widespread adoption of wearable devices in our daily lives, the demand and appeal for remote patient monitoring have significantly increased. Most research in this field has concentrated on collecting sensor data, visualizing it, and analyzing it to detect anomalies in specific diseases such as diabetes, heart disease and depression. However, this domain has a notable gap in the aspect of human-machine interaction. This paper proposes REMONI, an autonomous REmote health MONItoring system that integrates multimodal large language models (MLLMs), the Internet of Things (IoT), and wearable devices. The system automatically and continuously collects vital signs, accelerometer data from a special wearable (such as a smartwatch), and visual data in patient video clips collected from cameras. This data is processed by an anomaly detection module, which includes a fall detection model and algorithms to identify and alert caregivers of the patient's emergency conditions. A distinctive feature of our proposed system is the natural language processing component, developed with MLLMs capable of detecting and recognizing a patient's activity and emotion while responding to healthcare worker’s inquiries. Additionally, prompt engineering is employed to integrate all patient information seamlessly. As a result, doctors and nurses can access real-time vital signs and the patient's current state and mood by interacting with an intelligent agent through a user-friendly web application. Our experiments demonstrate that our system is implementable and scalable for real-life scenarios, potentially reducing the workload of medical professionals and healthcare costs. A full-fledged prototype illustrating the functionalities of the system has been developed and being tested to demonstrate the robustness of its various capabilities.
\end{abstract}


\section{Introduction} 
The imbalance between the number of patients and medical professionals has always been a significant challenge for the healthcare system. While the population is rapidly increasing, the workforce of medical professionals is growing slowly due to the high requirements of this field. This leads to a high and stressful workload for doctors and nurses, which could affect the quality of care services.

Since the development of the Internet of Things (IoT), many studies have tried to apply it to the healthcare environment to support doctors and nurses with their daily tasks, such as systematizing medical records~\cite{healthrecord}, automating medical appointment booking~\cite{iot_booking}, and moving essential medical examinations to telehealth~\cite{telehealth}. Recently, large language models (LLMs) have been experiencing significant growth. Numerous LLMs related to healthcare have been developed~\cite{clinicalt5,meditron,medpalm,medalpaca} to improve human-machine interaction, enabling them to respond to medical inquiries from patients or questions in medical textbooks. However, to date, there has been limited research on creating virtual assistants for medical professionals.

\begin{figure*}[ht]
\centering
\includegraphics[width=2\columnwidth]{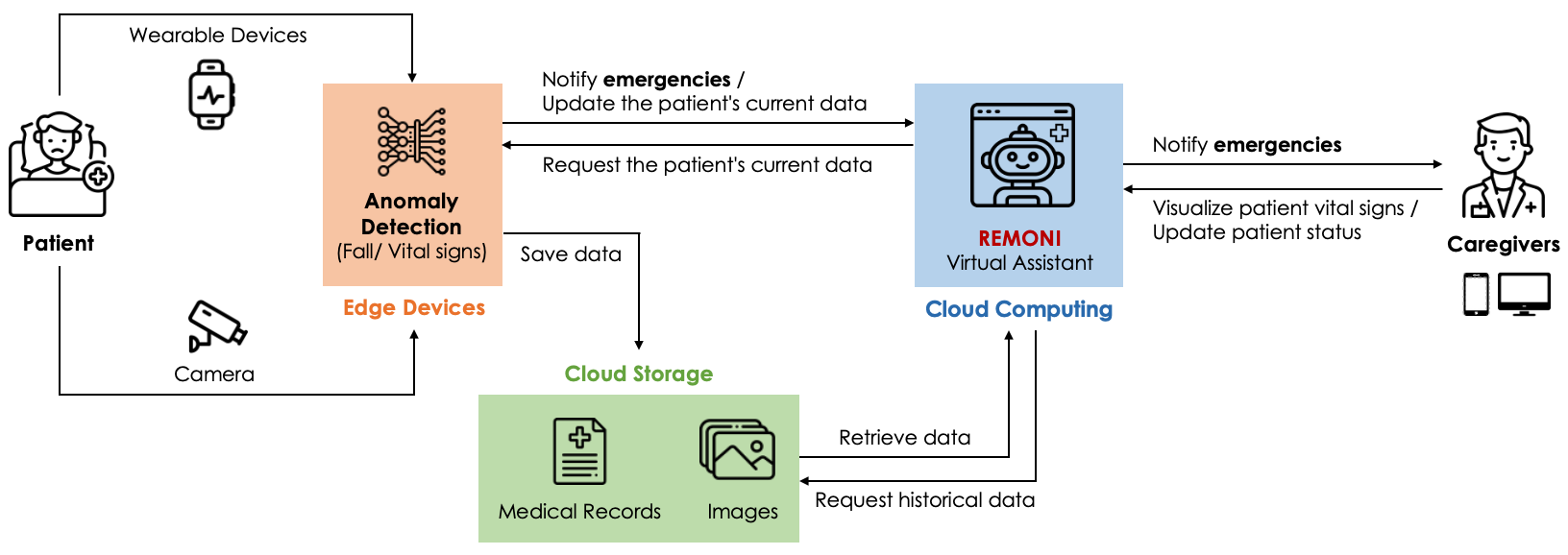}
\caption{Proposed Internet of Things System.}
\label{fig:proposed-model-architecture}
\end{figure*}

This paper proposes REMONI, an autonomous REmote health MONItoring system. Our system has two main components: an anomaly detection module and a smart natural language processing component capable of recognizing patient status (activity and emotion). These two components are built with a scalable and implementable IoT system to operate in practice. The main functions of the system are to promptly report the patient's abnormalities to the doctor, remotely monitor the patient's health, and support the doctor in retrieving patient data.

Our contributions are as follows:
\begin{itemize}
    \item To the best of our knowledge, this is the first work for the concept of an all-in-one virtual assistant for medical professionals, which automatically gathers data from sensors, stores it, detects anomalies, and uses LLMs to converse with doctors.
    
    \item We propose an IoT system architecture that connects sensors, wearable devices, LLMs, and end-user web applications using edge devices, cloud storage, and cloud computing. The architecture is flexible and easy to scale to a larger number of sensors and users, like in a hospital system.
    
    \item We demonstrate the medical virtual assistant's effectiveness and potential in improving the telehealth field.
    
\end{itemize}

\section{Related Work}\label{sec:related-work}
For a long time, researchers have been keen on developing virtual assistants for the healthcare sector to enhance human-machine interaction and improve care service quality.

During the early stages of Computer Vision development with Convolutional Neural Networks (CNNs), and when Natural Language Processing (NLP) was still in its infancy, Yoon et al.~\cite{Park-2019} created a projection-based augmented reality system to assist the elderly daily. They developed a camera capable of panning and tilting to monitor a wide area. Additionally, they utilized some deep learning frameworks for posture estimation, facial recognition, and object detection. With these capabilities, they constructed a system to project images onto surfaces (walls or tables) for the elderly to interact with. This system enabled the elderly to use their phones to see who was at the door and decide whether to grant entry. It also displayed daily reminders for taking medications or closing doors.

As NLP advanced, Nikita et al.~\cite{medical-chatbot} proposed a chatbot that employed machine learning techniques with metrics like TF-IDF, Stemming, n-grams, and cosine similarity to analyze users' healthcare-related queries. They aimed to create a chatbot to offer patients preliminary advice before consulting a doctor.

In the past two years, with the rapid development of LLMs, several medical chatbots have been created~\cite{clinicalt5,meditron,medpalm,medalpaca}.These chatbots primarily focus on answering patients' medical questions. Yunxiang et al.~\cite{chatdoctor} constructed a dataset of authentic patient-doctor conversations, gathering around 100k interactions from the online medical consultation website HealthCareMagic and an additional 10k conversations from another online medical consultation site, iCliniq.

While there is significant research in this field, it is evident that most efforts are centered on the patient side. A noticeable gap exists in developing virtual assistants specifically designed for medical professionals.

\section{Proposed Framework}\label{sec:proposed-work}
The framework proposed in this study comprises three modules: an Anomaly Detection module, a Natural Language Processing Engine, and an Internet of Things module for system deployment. Each of these is described in details below.

\subsection{Anomaly Detection}
Within our system, anomaly detection is a crucial process for recognizing irregular patterns that may indicate falls or critical changes in the patient's health status.
\subsubsection{FallAllD Dataset for Fall Detection}
\label{sec:FallAllD-Dataset}
\begin{figure}[htb]
\centering
\includegraphics[width=0.75\linewidth]{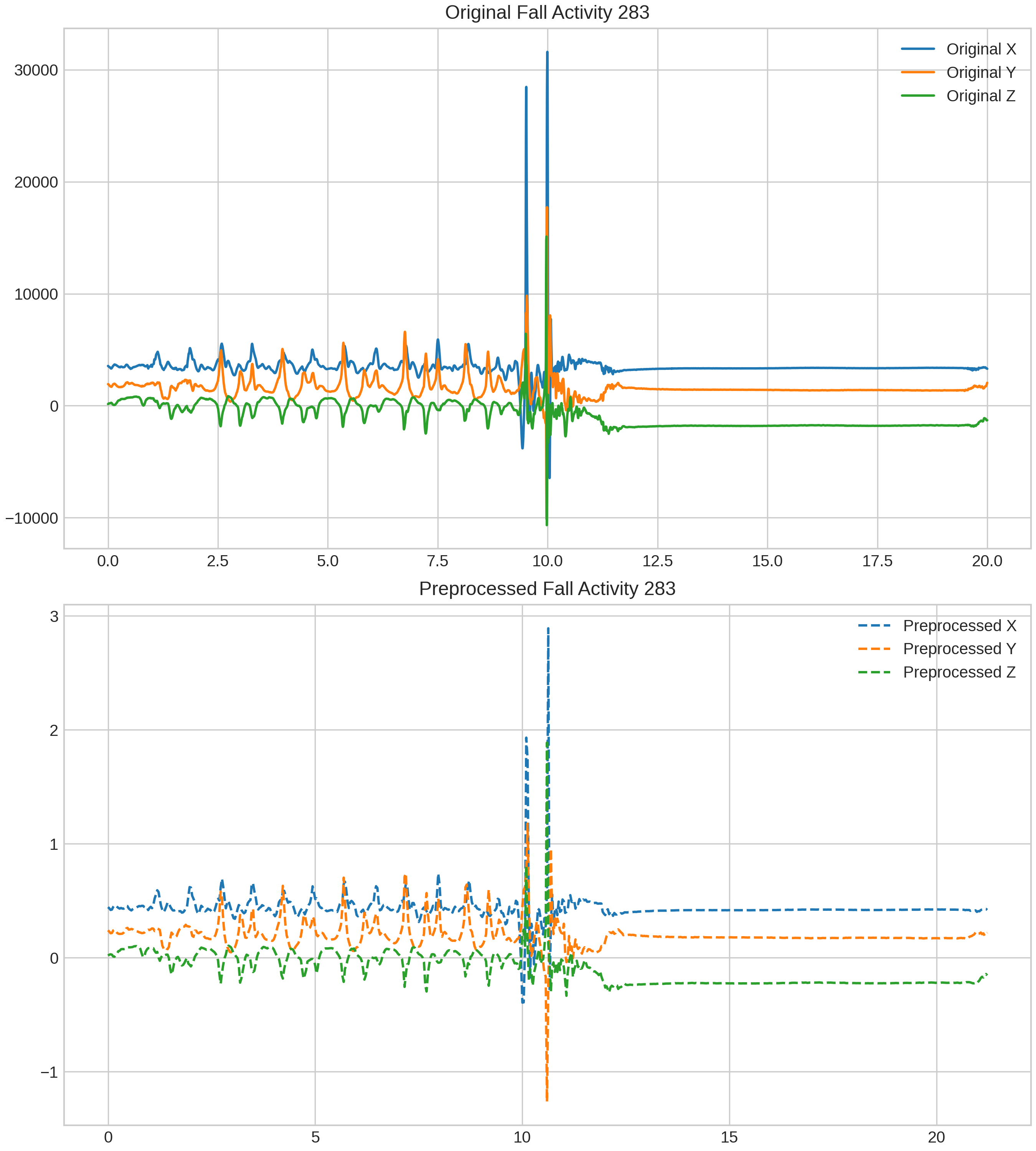}
\caption{Comparison of Original and Preprocessed Fall Activity from FallAllD Dataset.}
\label{fig:fall_comparison}
\end{figure}

\begin{itemize} [leftmargin=*]
    \item \textbf{Dataset Description:} This study uses the FallAllD dataset~\cite{Saleh-2021}, capturing 35 human falls and 44 daily activities, collected with three data loggers in an outdoor environment, where the subjects fell on grass instead of on soft mats. Each data-logger is equipped with the inertial module LSM9DS1, containing a 3-axial accelerometer configured with a sampling frequency of \SI{238}{\Hz} and a measurement range of $\pm$\SI{8}{g}. Fifteen participants wore these devices simultaneously on their necks, wrists, and waists to comprehensively capture their motion signals. The data collection protocol covers various fall types, including scenarios followed by recovery, and considers different postures and causes such as slips, trips, and syncope. This inclusivity is crucial for developing fall detection systems that are robust and effective across diverse real-world situations.
    
    \item \textbf{Dataset Preprocessing:} 
    In preprocessing the FallAllD dataset, we focused on the accelerometer data obtained from the subjects' wrists. This choice is supported by the ability of accelerometer data to accurately capture motion patterns and dynamics critical to identifying falls, distinguishing them from activities of daily living (ADL) with high reliability. The literature further validates our approach as many researchers acknowledge the sufficiency of accelerometer data for effective fall detection~\cite{Liu-2022, Liu-2023,nguyen-2023}.  
    Furthermore, we rescaled the original FallAllD dataset's accelerometer data from $\pm$\SI{8}{g} to $\pm$\SI{1}{g} and downsampled it from \SI{238}{\Hz} to \SI{32}{\Hz}. Fig.~\ref{fig:fall_comparison} compares the original FallAllD dataset and the preprocessed data for one random fall activity. It’s evident that despite the reduced sampling rate and the range, the peaks associated with the fall are still clear in the preprocessed data, which indicates that the preprocessing retains the critical features necessary for fall detection.
\end{itemize}

\subsubsection{Proposed Hybrid Deep Learning Model}
We build upon the foundation of our previously proposed hybrid deep learning (HDL) model~\cite{kharrat-2023}, which combines the strengths of convolutional neural networks (CNNs) for spatial feature extraction and long short-term memory (LSTM) networks for temporal sequence analysis, to address the time series challenge of human activity and fall detection. Retaining the model's core architecture, we have implemented minor refinements to enhance its performance further. The revised model, employing a sigmoid activation function in the final layer for optimal binary classification and compiled with an Adam optimizer and binary cross-entropy loss, offers enhanced performance and reliability in our real-world application.

\subsubsection{Threshold-Based Assessment of Vital Sign Anomalies}
In addition to the fall detection capability, our system integrates a threshold-based algorithm to detect anomalies in vital signs and automatically alert caregivers in case of health risks or emergencies. This Python model employs predefined healthy ranges for five key vital signs, namely body temperature ($36.5$--$37.2^\circ\mathrm{C}$), heart rate ($60$--$100$ beats per minute), respiration rate ($12$--$20$ breaths per minute at rest), blood pressure ($90/60$ to $120/80$ mmHg)~\cite{Vital-Signs}, and oxygen saturation (SpO$_2 \geq 95\%$)~\cite{Oxygen-Saturation}. REMONI continuously monitors these vital signs and upon detecting any values outside the healthy range, it promptly sends a notification alerting the medical personnel. 

\subsection{Natural Language Processing Engine}
\begin{figure*}[ht]
\centering
\includegraphics[width=2\columnwidth]{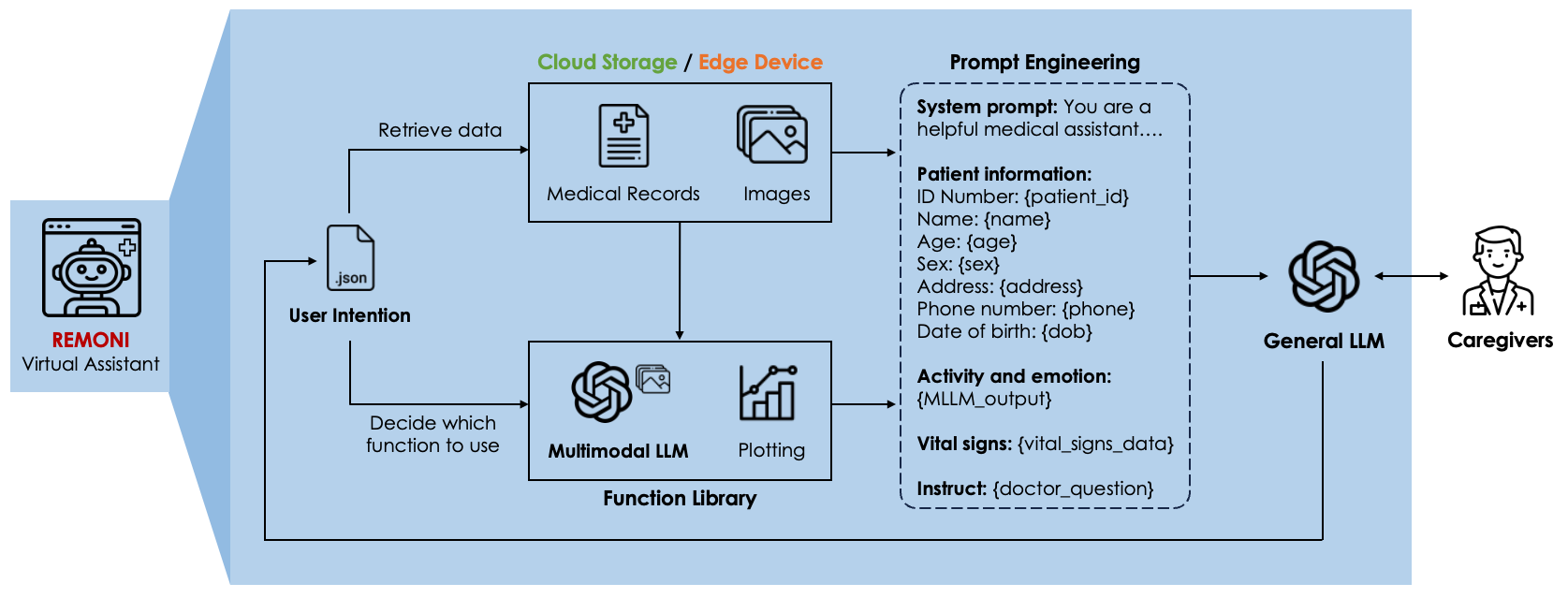}
\caption{Natural Language Processing Engine.}
\label{fig:proposed_nlp_engine}
\end{figure*}

In the system, medical professionals communicate directly with REMONI, a virtual assistant, through a web application. REMONI is powered by an NLP engine, as shown in Fig.~\ref{fig:proposed_nlp_engine}, which operates in three main stages: intention detection, data preparation, and final output production.

Initially, caregiver inquiries are processed by a General Large Language Model (LLM) to discern the user's intent. A system prompt, crafted to detail the intent detection task, is supplied to the General LLM alongside the doctor's question. This enables the General LLM to identify and output the user's intention in JSON format with the following keys: $patient\_id$, $list\_date$, $list\_time$, $vital\_sign$, $is\_plot$, $is\_recognition$, and $is\_image$.

In the second stage, the engine utilizes the $list\_date$ and $list\_time$ keys to determine whether to retrieve necessary data from cloud storage or the edge device. The $is\_plot$ and $is\_recognition$ keys indicate whether a plotting function or a MLLM is required to support the final response. The MLLM, equipped with its system prompt, is tasked with describing the patient's current status in an image, focusing on activity and emotion, which are crucial for depicting the patient's condition.

Finally, after collating all the information, the agent compiles it into an endpoint prompt that includes the patient's personal information, activity and emotion (as identified by the MLLM), vital signs (sourced from medical records in cloud storage or the edge device), and the user's question as instructions. A system prompt at the beginning of the endpoint prompt instructs the General LLM to use the data in the prompt to answer the user's questions accurately.

In summary, the NLP engine employs a General LLM for two distinct tasks (intent detection and final output production), each with a specific system prompt to guide the LLM's output according to the task. Additionally, the engine features a function library that includes a MLLM for recognizing the patient's activity and emotion and a plot function for visualizing the vital sign data.

\subsection{Internet of Things System}

The proposed IoT system comprises four main components: sensors, edge devices, cloud storage, and cloud computing. An overview of the architecture is illustrated in Fig.~\ref{fig:proposed-model-architecture}. Currently, our system is compatible with smartwatches and cameras as sensors.

At the core of our deployment lies the seamless integration of wearable technology. A custom wearable application serves as the conduit for data acquisition, harnessing the watch's sensors to collect accelerometer and physiological data. This application uses the Wi-Fi API to transmit the collected data streams to a designated Python module hosted on an edge device for further processing.

The edge device collects real-time data (accelerometer readings, vital signs, and visual information) from these sensors. It processes it through anomaly detection models (for falls and vital signs) to quickly identify emergencies. The device immediately sends an alert to the cloud computing platform if an emergency is detected. Otherwise, it periodically uploads vital signs and visual data to cloud storage for preservation and future use. Cloud storage archives historical data and responds to requests from the NLP engine for the data required to address user queries.

Regarding cloud computing, it serves as the host for the web application. Its primary role is to receive emergency alerts from the edge device and notify medical professionals. Additionally, it facilitates the NLP engine's communication with the edge device for current data and with cloud storage for historical data.

\section{Experiments}\label{sec:experiments}
\subsection{Deployment}\label{sec:deployment}

The implementation of REMONI involves a carefully orchestrated setup of interconnected components, all operating concurrently on the edge device, each contributing to the system's robustness and effectiveness in remote health monitoring.

The project utilizes the ACER Nitro AN515 as the edge device, which is equipped with an i5 9th-gen CPU, a GTX 1660 TI GPU, 8 GB of RAM, and 1 TB SSD storage. For imaging, we selected the Logitech 720p HD Webcam for its image quality and ease of use.

For data acquisition, we deployed a Tizen wearable web app specifically designed for the Samsung Galaxy Watch 3 (4GB of RAM, 380 mAh battery, multiple physiological sensors), chosen for its advanced sensors, reliability, and seamless integration with our system. AWS S3 and AWS EC2 were selected for their stable performance for cloud storage and computing.

These intricately integrated components form a robust deployment framework that delivers scalable, reliable, and responsive remote health monitoring capabilities to caregivers and healthcare professionals.

\subsection{Results and Discussion}\label{sec:results}

To assess the system's effectiveness, we examine the performance of the fall detection model, which utilizes accelerometer data from the Samsung Galaxy Watch 3. Additionally, we evaluate the accuracy of activity and emotion recognition using MLLMs and analyze the implementation time required for various tasks within the system.

\subsubsection{Fall detection model}

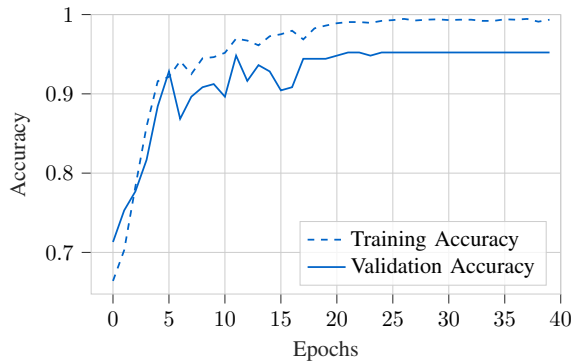
\begin{figure}[!ht]
\centering
\resizebox{0.9\columnwidth}{!}{
\begin{tikzpicture}

\definecolor{customblue}{RGB}{0, 100, 200}
\definecolor{dimgray85}{RGB}{85,85,85}
\definecolor{gainsboro229}{RGB}{229,229,229}
\definecolor{lightgray204}{RGB}{204,204,204}
\definecolor{mediumseagreen85168104}{RGB}{85,168,104}

\definecolor{darkslategray38}{RGB}{38,38,38}
\definecolor{indianred1967882}{RGB}{196,78,82}

\begin{groupplot}[group style={group size=1 by 2}, width=9cm, height=6cm]
\nextgroupplot[
axis line style={lightgray204},
legend cell align={left},
legend style={
  fill opacity=0.8,
  draw opacity=1,
  text opacity=1,
  at={(0.97,0.03)},
  anchor=south east,
  draw=lightgray204,
},
tick align=outside,
tick pos=left,
x grid style={lightgray204},
xlabel=\textcolor{darkslategray38}{Epochs},
xmajorgrids,
xmin=-1.95, xmax=40,
xtick style={color=darkslategray38},
y grid style={lightgray204},
ylabel=\textcolor{darkslategray38}{Accuracy},
ymajorgrids,
ymin=0.647490042448044, ymax=1,
ytick style={color=darkslategray38}
]
\addplot [thick, customblue, dashed]
table {%
0 0.664015889167786
1 0.702783286571503
2 0.782803177833557
3 0.85884690284729
4 0.916004002094269
5 0.921471178531647
6 0.941351890563965
7 0.924950301647186
8 0.944831013679504
9 0.946322083473206
10 0.951789259910583
11 0.969681918621063
12 0.967196822166443
13 0.961232602596283
14 0.972663998603821
15 0.975149095058441
16 0.9796222448349
17 0.968687891960144
18 0.982604384422302
19 0.986083507537842
20 0.989065587520599
21 0.990556657314301
22 0.990556657314301
23 0.989562630653381
24 0.992047727108002
25 0.993041753768921
26 0.994532823562622
27 0.992544710636139
28 0.993538796901703
29 0.99403578042984
30 0.993041753768921
31 0.993538796901703
32 0.993538796901703
33 0.992047727108002
34 0.992047727108002
35 0.99403578042984
36 0.993538796901703
37 0.994532823562622
38 0.991053700447083
39 0.993538796901703
};
\addlegendentry{Training Accuracy}
\addplot [thick, customblue]
table {%
0 0.713147401809692
1 0.752988040447235
2 0.776892423629761
3 0.816733062267303
4 0.884462177753448
5 0.928286850452423
6 0.868525922298431
7 0.896414339542389
8 0.908366560935974
9 0.912350594997406
10 0.896414339542389
11 0.948207199573517
12 0.916334688663483
13 0.936254978179932
14 0.928286850452423
15 0.904382467269897
16 0.908366560935974
17 0.94422310590744
18 0.94422310590744
19 0.94422310590744
20 0.948207199573517
21 0.952191233634949
22 0.952191233634949
23 0.948207199573517
24 0.952191233634949
25 0.952191233634949
26 0.952191233634949
27 0.952191233634949
28 0.952191233634949
29 0.952191233634949
30 0.952191233634949
31 0.952191233634949
32 0.952191233634949
33 0.952191233634949
34 0.952191233634949
35 0.952191233634949
36 0.952191233634949
37 0.952191233634949
38 0.952191233634949
39 0.952191233634949
};
\addlegendentry{Validation Accuracy}
\end{groupplot}

\end{tikzpicture}
}
\caption{CNN-LSTM Traning vs. Validation Accuracy on the Preprocessed FallAllD dataset.}
\label{fig:Preprocessed_FallAllD_Accuracy}
\end{figure}

Fig.~\ref{fig:Preprocessed_FallAllD_Accuracy} illustrates the training and validation accuracy over epochs. The model quickly reaches a high level of accuracy during training, which is consistently maintained throughout the validation phase. The convergence of training and validation accuracy suggests that the model generalizes well and exhibits minimal overfitting.

\begin{figure}[!ht]
\centering
\resizebox{0.9\columnwidth}{!}{
  \resizebox{\columnwidth}{!}{%
\begin{tikzpicture}

\definecolor{dimgray85}{RGB}{85,85,85}
\definecolor{gainsboro229}{RGB}{229,229,229}
\definecolor{ghostwhite247251255}{RGB}{247,251,255}
\definecolor{midnightblue848107}{RGB}{8,48,107}
\definecolor{darkslategray38}{RGB}{38,38,38}
\definecolor{lightgray204}{RGB}{204,204,204}

\begin{axis}[
axis background/.style={fill=gainsboro229},
axis line style={white},
colorbar,
colorbar style={ylabel={}},
colormap={mymap}{[1pt]
  rgb(0pt)=(0.968627450980392,0.984313725490196,1);
  rgb(1pt)=(0.870588235294118,0.92156862745098,0.968627450980392);
  rgb(2pt)=(0.776470588235294,0.858823529411765,0.937254901960784);
  rgb(3pt)=(0.619607843137255,0.792156862745098,0.882352941176471);
  rgb(4pt)=(0.419607843137255,0.682352941176471,0.83921568627451);
  rgb(5pt)=(0.258823529411765,0.572549019607843,0.776470588235294);
  rgb(6pt)=(0.129411764705882,0.443137254901961,0.709803921568627);
  rgb(7pt)=(0.0313725490196078,0.317647058823529,0.611764705882353);
  rgb(8pt)=(0.0313725490196078,0.188235294117647,0.419607843137255)
},
point meta max=1.00,
point meta min=0.00,
tick align=outside,
tick pos=left,
x grid style={white},
xlabel=\textcolor{darkslategray38}{Predicted labels},
xmin=0, xmax=2,
xtick style={color=darkslategray38},
xtick={0.5,1.5},
xticklabels={{\footnotesize Non Fall},{\footnotesize Fall}},
y dir=reverse,
y grid style={lightgray204},
ylabel=\textcolor{darkslategray38}{True labels},
ymin=0, ymax=2,
ytick style={color=darkslategray38},
ytick={0.5,1.5},
yticklabel style={rotate=90.0},
yticklabels={{\footnotesize Non Fall},{\footnotesize Fall}},
]
\addplot graphics [includegraphics cmd=\pgfimage,xmin=0, xmax=2, ymin=2, ymax=0] {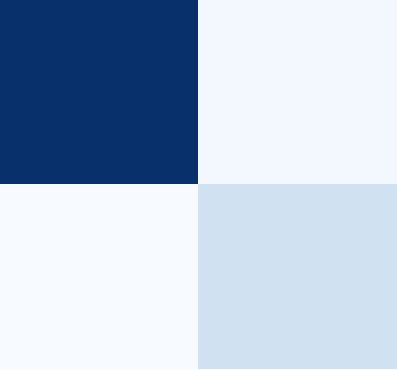};
\draw (axis cs:0.5,0.5) node[
  scale=0.90,
  text=white,
  rotate=0.0
]{0.99};
\draw (axis cs:1.5,0.5) node[
  scale=0.90,
  text=darkslategray38,
  rotate=0.0
]{0.01};
\draw (axis cs:0.5,1.5) node[
  scale=0.90,
  text=darkslategray38,
  rotate=0.0
]{0.02};
\draw (axis cs:1.5,1.5) node[
  scale=0.90,
  text=darkslategray38,
  rotate=0.0
]{0.98};
\end{axis}

\end{tikzpicture}
}
}
\caption{CNN-LSTM Model: Confusion Matrix for Test Data on the Preprocessed FallAllD Dataset.}
\label{fig:Preprocessed_FallAllD_confusion_matrix}
\end{figure}
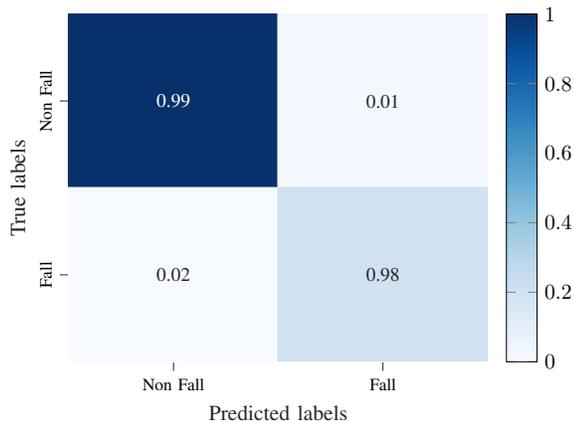

The confusion matrix in Fig.~\ref{fig:Preprocessed_FallAllD_confusion_matrix} displays the model's performance in classifying fall and non-fall events in the test data from the preprocessed FallAllD dataset. The high values on the matrix's diagonal (0.99 for Non-Fall, 0.98 for Fall) indicate a high true positive rate for both classes, demonstrating the model's accuracy. The low off-diagonal values (0.01 and 0.02) show a small fraction of misclassifications, confirming the model's reliability in fall detection.

Table~\ref{tab:FallAllD} provides a comparative analysis of various models tested on the FallAllD dataset, such as Light Gradient-Boosting Machine (LightGBM), Support Vector Machine (SVM), Coarse-fine CNN combined with Gated Recurrent Unit (GRU), and Late AFVF (feature-level fusion applied to Actual Fusion within Virtual Fusion).The proposed HDL model demonstrates superior performance with a precision of 99\%, an accuracy, recall, and F1-score all at 98\%. This highlights the model's high performance using solely wrist-acquired accelerometer data, underscoring its potential for reliable and accurate fall detection in real-world applications.

\begin{table}[!ht]
\centering
\begin{threeparttable}
\caption{Benchmarking for the FallAllD dataset}
\label{tab:FallAllD}
\begin{tabular}
{|m{0.25\columnwidth}|m{0.23\columnwidth}|m{0.05\columnwidth}|m{0.05\columnwidth}|m{0.05\columnwidth}|m{0.05\columnwidth}|}
\toprule
Model & Sensor/Location & A & P & R& F1 \\
\midrule
LightGBM~\cite{Kim-2023} & Acc+Gyro/Wrist & 95\% & N/A & 91\% & 91\% \\
SVM~\cite{Nunez-2024} & Acc+Gyro/Wrist & 93\% & N/A & 87\% & 93\% \\
CNN-GRU \cite{Liu-2023} & Acc/Waist & 98\% & 96\% & 93\% & 94\% \\
Late AFVF~\cite{nguyen-2023} & Acc/Waist+Wrist & N/A & N/A & N/A & 96\% \\
\textbf{Proposed model} & Acc/Wrist & \textbf{98\%}    &\textbf{99\%}&\textbf{98\%} &\textbf{98\%}\\
\bottomrule 
\end{tabular}
\begin{tablenotes}[para,flushleft]
\item[] Acc: Accelerometer, Gyro: Gyroscope, A: Accuracy, P: Precision, R: Recall, F1: F1-score
\end{tablenotes}
\end{threeparttable}
\end{table} 

\begin{table}[htb]
\centering
\caption{Performance of MLLMs in Activity and Emotion Recognition}
\label{tab:multimodal}
\begin{tabular}{|l|c|c|c|c|}
 \toprule
 Model&Accuracy&Precision&Recall&F1-score\\
 \midrule
 LLaVA-Activity & 10\% & 22\% & 10\% & 7\% \\
 Video-ChatGPT-Activity  & 35\% & 49\% & 35\% &  33\% \\
 \textbf{GPT4-Vision-Activity}   & \textbf{51\%} & \textbf{58\%} & \textbf{51\%} &  \textbf{45\%} \\
\midrule
 LLaVA-Emotion & 15\% & \textbf{39\%} & 15\% &   12\% \\
 Video-ChatGPT-Emotion & 20\%  & \textbf{39\%} & 20\% &  19\% \\
 \textbf{GPT4-Vision-Emotion}   & \textbf{41\%}    &35\% & \textbf{41\%} &  \textbf{31\%} \\

 \bottomrule 
\end{tabular}
\end{table}

\subsubsection{Activity and Emotion Recognition}
To assess the system's capability to recognize human activity and emotion, we selected the HACER dataset~\cite{hacer} for its high relevance to the project. The dataset comprises around 500 video clips featuring ten participants, including men and women. Each video is labelled with both activity and emotion categories. The dataset features nine activity classes: $drinking$, $putting\_on\_glasses$, $putting\_on\_jacket$, $reading$, $sitting\_down$, $standing\_up$, $taking\_off\_glasses$, $taking\_off\_jacket$, and $writing$. Additionally, it includes five emotion classes: $angry$, $disgust$, $happy$, $neutral$, and $sad$.

We crafted a prompt similar to the system prompt used by the MLLM in our system. The difference is that we explicitly list the activity and emotion classes in the system prompt and ask the models to identify the correct one. If no person is visible in the image, the models should respond with `unidentifiable' for both activity and emotion. Similarly, if the person's face is unclear, particularly in surveillance footage, the models should output `unidentifiable' for the emotion while still attempting to identify the activity.

For evaluation, we utilized three multimodal large language models: LLaVA~\cite{llava}, Video-ChatGPT~\cite{videochatgpt}, and GPT4-Vision~\cite{gpt4}. LLaVA and GPT4-Vision analyze a single frame, specifically the middle frame extracted from the video. In contrast, Video-ChatGPT processes approximately 100 frames. 

As shown in Table~\ref{tab:multimodal}, LLaVA had the lowest performance, with 10\% accuracy for activity and 15\% for emotion. This may be due to its reliance solely on image input. Additionally, it frequently classifies activities and emotions as `unidentifiable', suggesting a lack of focus on human faces or poses for activity and emotion detection.

While the training process of Video-ChatGPT enriched the dataset's context, enabling it to provide more detailed responses, it ranked second with an accuracy of 35\% for activity recognition and 20\% for emotion detection. The model's propensity for lengthy answers sometimes resulted in missing the correct classes, which did not align with our need for straightforward recognition of emotions and activities.


GPT4-Vision showed the best overall performance, with 51\% accuracy for activity and 41\% for emotion, despite only using the middle frame of the video. Its performance significantly surpasses that of LLaVA. However, a closer examination of the precision values for emotion reveals that GPT4-Vision tends to output `happy' and `neutral' as the emotion class, resulting in slightly lower precision. In contrast, despite their lower accuracy, the other two models produce outputs that span across various emotion classes.

Although GPT4-Vision scores the highest in most metrics, its performance is still relatively low. This may be attributed to our stringent criteria for determining accurate classifications. For instance, if a person is sitting and drinking water, the dataset classifies it as drinking, but if the models detect it as sitting, it is considered a false detection. We are in the process of developing a more equitable evaluation method. Additionally, the system is set to be implemented in a real-world scenario shortly and will undergo further fine-tuning with actual data from a clinic in Abu Dhabi. As a result, we anticipate an improvement in the performance of the MLLMs. For the demo phase, GPT4-Vision has been selected as the MLLM for the NLP engine.

\subsubsection{Response Time Delays}
The overall system latency was evaluated through three main approaches, each encompassing various questions based on the data requested. Each scenario's time is measured ten times before calculating the average and variance.

Firstly, as shown in Table~\ref{tab:response_time}, the response time to queries about real-time (instant) data, such as the patient's current state or vital signs, is assessed. . These questions necessitate communication between the cloud computing system and the edge device to retrieve the latest sensor data and provide answers. Moreover, the latency in responding to requests for historical information stored in the cloud database is also investigated.

\begin{table}[htb]
\centering
\caption{Response time analysis of the REMONI system for various question types}
\label{tab:response_time}
\begin{tabular}{|c|c|c|}
\toprule
Question Type \textbackslash Data type &  Instant Data  & Historical Data \\
\midrule
\textbf{Image Retrieval} & $ 6.84s \pm 2.09s$ & $ 3.37s \pm 0.15s$ \\
\textbf{Vital Sign Retrieval} & $ 5.69s \pm 0.06s$ & $ 6.62s \pm 8.73s$\\
\textbf{Image Description} & $ 9.56s \pm 6.11s$ & $ 9.75s \pm 9.70s$ \\
\textbf{Plotting Vital Sign} & N/A & $ 4.61s \pm 1.58s$ \\
\bottomrule
\end{tabular}
\end{table}

Overall, the system's response time to user queries is under 20 seconds. Questions requiring the NLP engine to use MLLM to determine the patient's current state (activity and emotion) take longer to answer. Additionally, retrieving instant image data from edge devices takes much longer than accessing historical image data from cloud storage.

Lastly, the system's responsiveness in sending emergency alerts from the edge device to the caregiver was examined, with response times averaging around 0.2 seconds.

\section{Conclusion}\label{sec:conclusion}

Our proposed system, REMONI, is an autonomous system that enhances human-machine interaction in REmote health MONItoring. Utilizing MLLMs, IoT, and wearable devices, REMONI provides a comprehensive solution for the continuous and automatic collection of vital signs, accelerometer data, and visual data. It also facilitates anomaly detection and seamless communication between medical professionals and the system. The system is designed to be easily scalable to larger environments involving multiple medical professionals and many patients. Additionally, more wearable devices and anomaly detection algorithms can be quickly integrated into our proposed IoT system architecture. Our experiments demonstrate the system's feasibility for real-life applications. We believe in its potential to alleviate the workload of medical professionals and reduce healthcare costs. In future work, we plan to deploy the system in real-life settings and further enhance the performance of its components.

%
\bibliographystyle{IEEEtran}
\bibliography{References/references}

\end{document}